\documentclass{article}

\usepackage{PRIMEarxiv}
\usepackage{amsmath}
\usepackage[utf8]{inputenc} % allow utf-8 input
\usepackage[T1]{fontenc}    % use 8-bit T1 fonts
\usepackage{hyperref}       % hyperlinks
\usepackage{url}            % simple URL typesetting
\usepackage{booktabs}       % professional-quality tables
\usepackage{amsfonts}       % blackboard math symbols
\usepackage{nicefrac}       % compact symbols for 1/2, etc.
\usepackage{microtype}      % microtypography
\usepackage{lipsum}
\usepackage{fancyhdr}       % header
\usepackage{graphicx}       % graphics
\graphicspath{{media/}}     % organize your images and other figures under media/ folder

\title{FedRobo: Federated Learning Driven Autonomous Inter Robots Communication For Optimal Chemical Sprays}

\author{
  Jannatul Ferdaus \\
  Computer Science and Engineering \\
  University of Asia Pacific \\
  Dhaka, Bangladesh\\
\texttt{syeda.jannat243@gmail.com} 
\\
\And
  Sameera Pisupati \\
  Computer Science \\
  University of North Texas \\
  TX, USA\\
\texttt{sameerapisupati@my.unt.edu}
\\
\And
  Mahedi Hasan \\
  Computer Science and Engineering \\
  Green University of Bangladesh \\
  Dhaka, Bangladesh\\
\texttt{mehedigub2020@gmail.com}
\\
\And
   Sathwick Paladugu \\
  School of Computing \\
  Southern Illinois University Carbondale, IL, USA\\
\texttt{sathwick.paladugu@siu.edu}
}

\begin{document}
\maketitle

\begin{abstract}
Federated Learning enables robots to learn from each other's experiences without relying on centralized data collection. Each robot independently maintains a model of crop conditions and chemical spray effectiveness, which is periodically shared with other robots in the fleet. A communication protocol is designed to optimize chemical spray applications by facilitating the exchange of information about crop conditions, weather, and other critical factors. The federated learning algorithm leverages this shared data to continuously refine the chemical spray strategy, reducing waste and improving crop yields. This approach has the potential to revolutionize the agriculture industry by offering a scalable and efficient solution for crop protection. However, significant challenges remain, including the development of a secure and robust communication protocol, the design of a federated learning algorithm that effectively integrates data from multiple sources, and ensuring the safety and reliability of autonomous robots. The proposed cluster-based federated learning approach also effectively reduces the computational load on the global server and minimizes communication overhead among clients.

%   We propose Federated learning X-ray weapon detection in airports using Machine Learning. To reduce manual inspection for weapon detection to some level by preserving data privacy. Training models in one airport and sharing across the model using federated learning could leverage a few of the newer problems. Modern airports have access to a lot of luggage X-ray data that may be used to train learning models, which can significantly reduce the manual inspection process. These Machine Learning models, for instance, can enhance image models and can choose appealing photos on their own. To the data center and training there using standard methods may be impossible due to the rich data's frequent privacy concerns, size, or both. We recommend an option that aggregates locally calculated updates to learn a shared model while leaving the training data scattered on the edge devices. On the other hand, by using federated learning, we only synchronize the model rather than the personal data. This process will ensure that we have data privacy concerns.
\end{abstract}

\section{Introduction}
\label{sec:introduction}
Federated learning is a type of machine learning in which multiple devices or agents work together to learn a common model without sharing raw data. Within the context of autonomy, federated learning can be used to improve system performance by inter-robot communication for optimal chemical sprays~\cite{yu2022towards, accurate2022crop}. 

\begin{figure}
\centering
\includegraphics[width=0.8\textwidth]{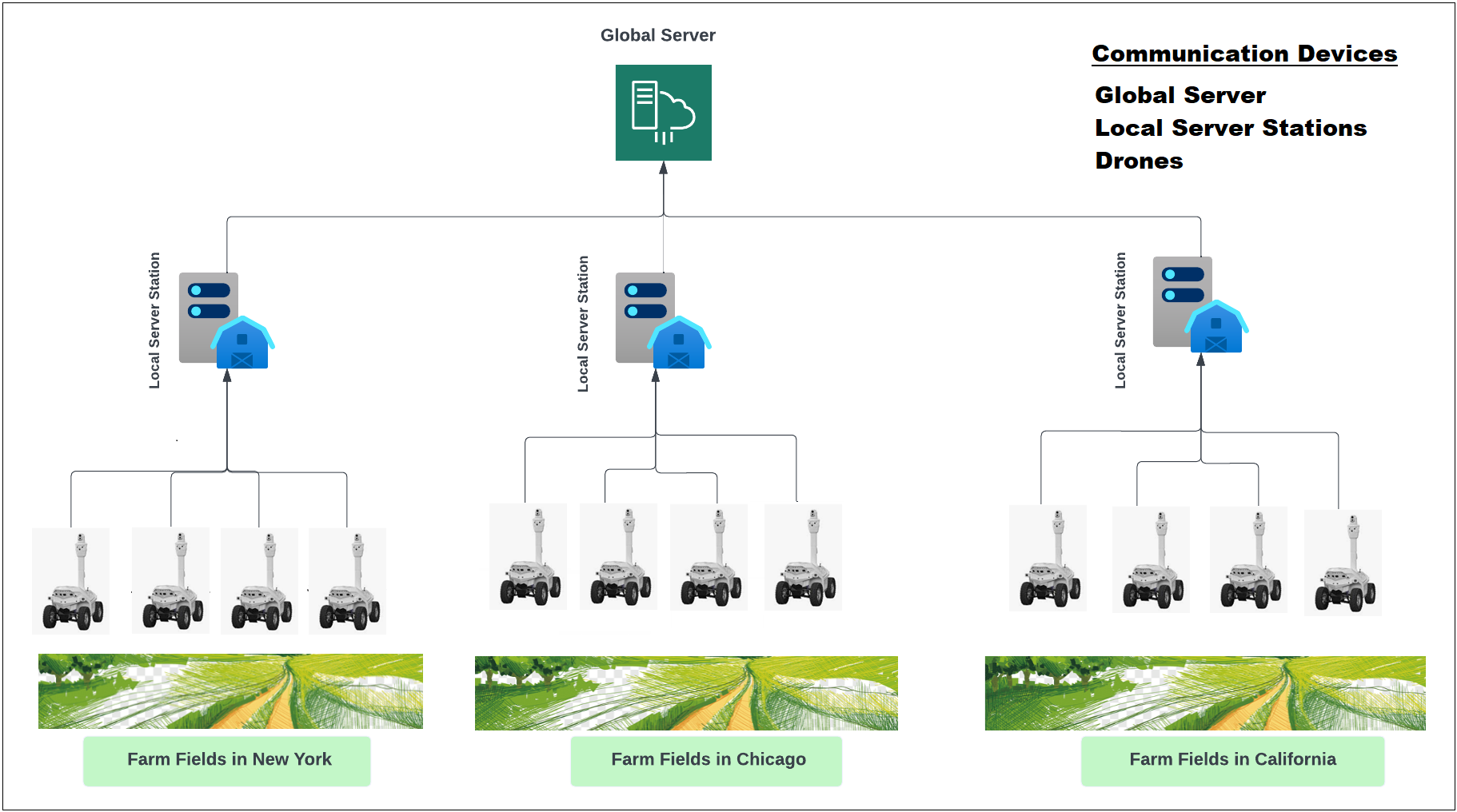}
\caption{ The architecture of autonomous robots communicating with the global server via local server station.} 
\label{fig:architecturelab}
\end{figure}

Autonomous inter-robot communication involves a group of robots communicating with each other to optimize their chemical spray patterns. The robots may have different sensors and spray systems and may need to adjust their spray patterns based on changes in the environment. Federated learning can help these robots learn from each other without sharing sensitive data. Figure~\ref{fig:architecturelab} shows a federated learning framework that consists of several local servers and one global server.

To implement federated learning in this context, each robot could have a local model that is trained on its own sensor data. The model updates are sent to the global server. The local models could then be aggregated into a global model that represents the collective knowledge of all the robots and the global server later pushes those updates to all the autonomous robots for awareness purpose. This global model could be used to optimize the spray patterns for all the robots. 

In this context, one advantage of federated learning is that it allows robots to adapt to changes in the environment without requiring them to constantly communicate with a central server. Instead, the robots can learn from each other and make local decisions based on the global model~\cite{yu2022towards}. However, there are also challenges associated with implementing federated learning in this context. These challenges include ensuring data privacy and security, dealing with communication delays and failures, and handling heterogeneous data and models.

In this scenario, autonomous robots equipped with chemical sprayers can collect data on the environment and the plants they are spraying. This data can include information such as plant type, growth stage, and weather conditions. Based on this data, the robots can use federated learning to collaborate on a shared model to optimize the chemical spraying process.

Each robot in the federated learning process trains a local model using its own data before sending updates to a central server. The central server aggregates the updates and returns to the robots a new global model, which they use for their next spraying task. This process can be repeated iteratively to continuously improve the spraying process's accuracy and efficiency~\cite{accurate2022crop}.

In this scenario, federated learning has several advantages. For starters, it allows the robots to collaborate on a shared model without sharing their data, which can aid in data privacy protection. Second, it enables the robots to continuously learn and adapt to changing environmental conditions and plant growth stages, which can improve the spraying process's effectiveness. Finally, it can help reduce the amount of chemicals used, which can benefit both the environment and the economy. However, there are some drawbacks to using federated learning in this context. One challenge is ensuring that the local models trained by the robots are representative of the entire population of plants being sprayed. Another challenge is ensuring that the updates sent by the robots to the central server are secure and cannot be intercepted or tampered with.

\textbf{Autonomous Robots.}
Autonomous robots are those that can complete tasks or make decisions without the need for human intervention. These robots can sense and perceive their surroundings, make decisions based on that data, and take actions to achieve their objectives~\cite{autonomous_mukhopadhyay_2007,autonomy_smithers_1997,intelligent_vidoni_2011,robots_kaminka_2007,introduction_correll_2015,autonomous_meystel_1991}.

Autonomous robots have numerous applications, including manufacturing, logistics, healthcare, agriculture, and exploration. Autonomous robots can be used in manufacturing for tasks such as assembly, inspection, and packaging. They can be used in logistics for tasks such as transportation, sorting, and delivery.
In healthcare, autonomous robots can be used for tasks such as patient monitoring, medication delivery, and disinfection.
They can be used in agriculture for tasks such as planting, harvesting, and crop health monitoring. They can be used in exploration for tasks such as mapping, surveying, and search and rescue.
Autonomous robots rely on a variety of technologies to
sense and perceive their environment, make decisions, and take~\cite{introduction_siegwart_2011,autonomous_meystel_1991,autonomous_kaminka_2012,robust_ferrell_1993,deliberation_ingrand_2017}
actions. These technologies include sensors, such as cameras
and lidar, that allow the robot to detect and interpret its
surroundings. They also include algorithms and software that allow the robot to make decisions based on that information, as well as actuators like motors and manipulators that allow the robot to perform actions.
However, there are some challenges to developing and deploying autonomous robots, such as ensuring safety and reliability, addressing ethical and legal concerns, and dealing with the potential impact on employment. Nonetheless, the potential benefits of autonomous robots in terms of increased efficiency, productivity, and safety make them an important research and development area ~\cite{sustainable_muzirafuti_2022,simulation_brugali_2014}.

\textbf{Federated Learning.}
Federated learning is a machine learning technique that enables multiple parties to collaboratively train a model without sharing their raw data. Instead of collecting all data in a central location, federated learning allows each participant to train the model locally on their own device or server. The updated model is then sent to a central server, which aggregates the changes and sends an updated model back to each participant. This process is repeated iteratively until the model converges to an acceptable level of accuracy.

Federated learning is useful in scenarios where data is sensitive or large, making it impractical to transfer it to a central server. By keeping data local, federated learning reduces privacy concerns and can improve training efficiency. Federated learning has been used in various applications, such as improving predictive text on mobile devices, improving healthcare research, and enabling collaborative machine learning in industrial IoT settings.

\section{Literature Review}
\label{sec:literature-review}
Federated learning has gained significant attention in recent years as a decentralized machine learning approach that allows multiple agents to collaboratively train a shared model without exchanging raw data. This technique is particularly useful in scenarios where data privacy and security are of paramount importance, such as in healthcare~\cite{talukder2022prediction,puppala2024flash,hossain2023collaborative, puppala2023machine}, social media~\cite{puppala2024socfedgpt} or security intense domains~\cite{puppala2022towards,talukder2022federated,talukder2022novel, puppala2024scale, talukder2022prediction}, and industrial IoT~\cite{nguyen2021federated}. The application of federated learning in autonomous systems, including robotics, has opened new avenues for enhancing the capabilities of such systems while ensuring data privacy and reducing communication overhead.

The integration of federated learning into robotic systems has been explored in various contexts, with the goal of improving the performance and autonomy of robots. For instance, Li et al.~\cite{li2020federated} demonstrated the use of federated learning in a swarm robotics scenario, where multiple robots collaboratively learned a model to optimize their collective behavior. This approach allowed the robots to improve their performance over time while minimizing the need for central coordination. Similarly, Kairouz et al.~\cite{kairouz2021advances} discussed the potential of federated learning to enable robots to learn from heterogeneous data sources, which is crucial in environments where robots are exposed to diverse conditions and tasks.

In the context of agricultural robotics, federated learning has been applied to optimize tasks such as chemical spraying and crop monitoring. Yu et al.~\cite{yu2022towards} proposed a federated learning framework for inter-robot communication, where multiple autonomous robots collaborated to optimize their chemical spray patterns without sharing raw sensor data. This approach not only improved the efficiency of the spraying process but also maintained data privacy, which is critical in agricultural settings. The study by Accurate et al.~\cite{accurate2022crop} further emphasized the benefits of federated learning in reducing chemical usage and adapting to changing environmental conditions.

While federated learning offers numerous advantages, its implementation in autonomous robotics poses several challenges. One major challenge is ensuring the security and integrity of the model updates exchanged between robots and the central server. As discussed by Bonawitz et al.~\cite{bonawitz2019towards}, securing federated learning involves protecting the communication channels and preventing adversarial attacks that could compromise the model. Moreover, the heterogeneity of data and models across different robots can lead to issues with model convergence and performance, as highlighted by McMahan et al.~\cite{mcmahan2017communication}.

Another challenge is the management of communication delays and failures, which can affect the synchronization of model updates and degrade the overall performance of the federated learning system. Strategies to address these challenges include the development of robust aggregation methods and the use of asynchronous communication protocols~\cite{li2019fedprox}. Additionally, ensuring that the local models trained by the robots are representative of the entire environment is crucial for the success of the federated learning process~\cite{wang2020federated}.

Future research in this area could focus on developing more efficient and secure federated learning algorithms tailored to the specific needs of autonomous robots. This includes addressing the challenges of model heterogeneity, improving communication efficiency, and enhancing the scalability of the learning process.

The application of federated learning in autonomous robots, particularly in the context of inter-robot communication for optimizing chemical sprays, holds great promise for improving the efficiency and effectiveness of these systems. While there are significant challenges to be addressed, ongoing research in this area continues to advance the state-of-the-art and pave the way for more autonomous and intelligent robotic systems.

\section{Methodology}
\label{sec:methodology}
Federated learning is a machine learning paradigm that enables multiple participants to collaboratively train a model without sharing their raw data. In this approach, the data remains on the device or at the edge, with the model training occurring locally.

The federated learning architecture typically includes the following components:

\textbf{Clients:} These are the devices or endpoints where the data resides. Each client holds a portion of the data used to train the machine learning model. In this context, each robot acts as a client, collecting data from the agricultural field and training a machine learning model locally. The trained model from each client (robot) is then sent to a central server for further processing.

\textbf{Server:} The central server plays a coordinating role, interacting with both global and local servers. It distributes the current model to each client, gathers model updates from them, and aggregates these updates to create a new global model. This aggregated model is then distributed to specific regions where multiple robots operate. Similarly, aggregated models from different regions are sent to the global server for further aggregation.

\textbf{Federated Learning Algorithm:} The algorithm used in federated learning is designed to accommodate the distributed nature of the data, which is stored on various devices. The algorithm operates through repeated training cycles, where each client trains the model using its local dataset, and these local models are averaged to produce a global model. This global model is then shared with all clients, enhancing their ability to predict new datasets.

\begin{figure}
\centering
\includegraphics[width=0.8\textwidth]{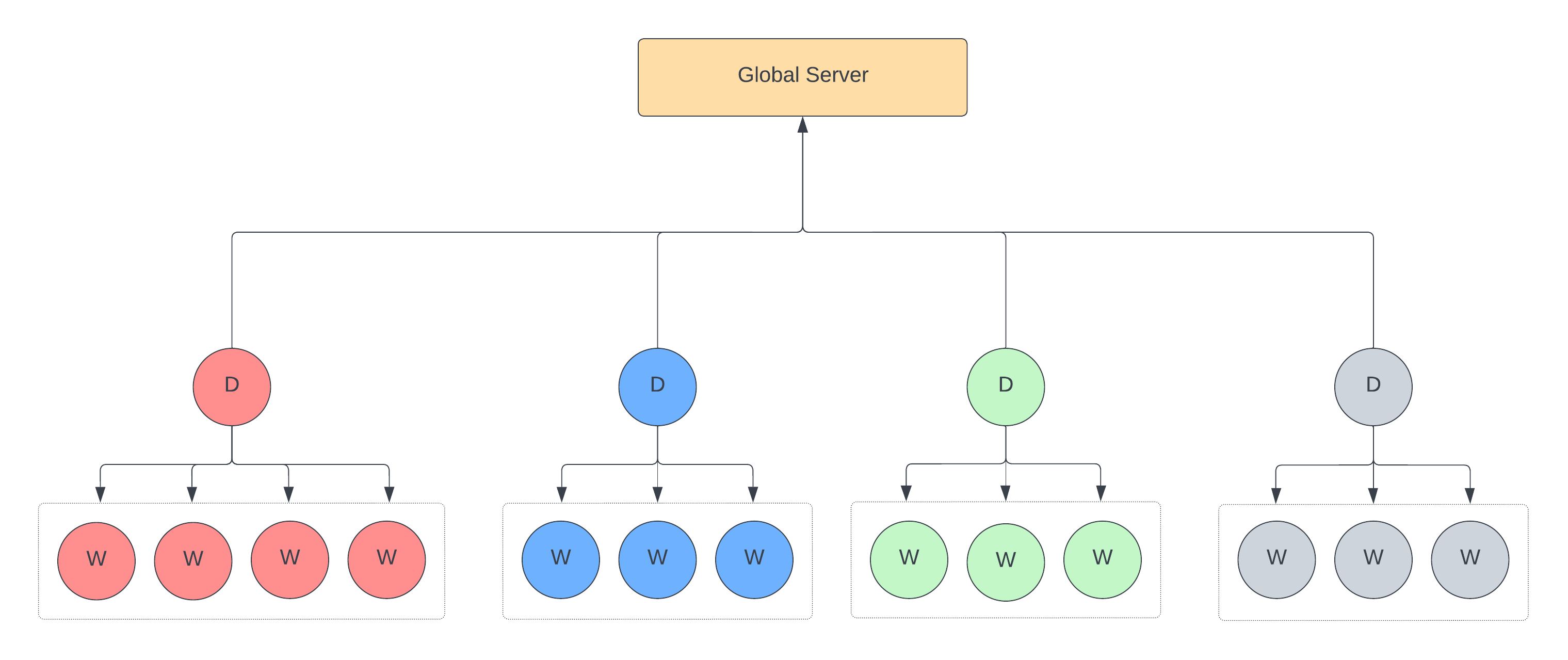}
\caption{An overview of federated learning architecture, where different colors represent different farm fields across the country.} 
\label{fig:systemarchitecture}
\end{figure}

Figures~\ref{fig:architecturelab} and ~\ref{fig:systemarchitecture} illustrate the federated learning framework. Both figures convey the same concept, with Figure~\ref{fig:systemarchitecture} providing a general overview of federated learning and Figure~\ref{fig:architecturelab} depicting robot communication within the federated learning framework. In Figure~\ref{fig:systemarchitecture}, the different colored circles labeled with `W' represent workers or clients, with color variations indicating the different areas where these workers operate. `D' represents the Drive node or server, which facilitates communication between the global server and the workers or clients. In Figure~\ref{fig:architecturelab}, the robots act as workers or clients, with each set of robots connected to a dedicated server for a specific area, which in turn communicates with the global server.

\textbf{Security and Privacy Mechanisms:} Federated learning relies on robust security and privacy mechanisms to protect the data of each client. These mechanisms include encryption, access control, and anonymization to ensure that client data remains secure throughout the process.

\begin{figure}
\centering
\includegraphics[width=0.9\textwidth]{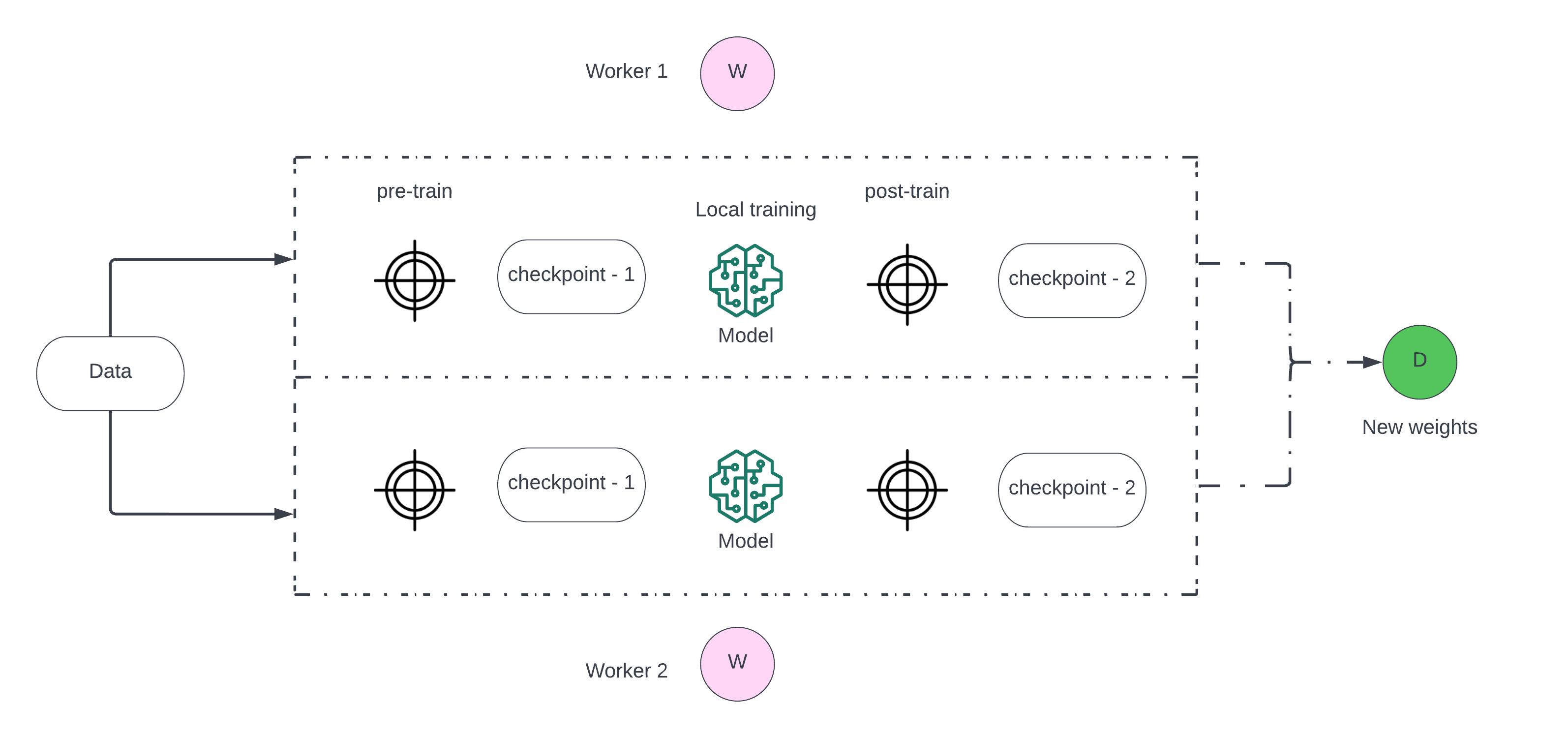}
\caption{Architecture showing how checkpointing is performed based on two model updates for two different nodes or autonomous robots.} 
\label{fig:fedsrccheckpoint}
\end{figure}

\textbf{Checkpointing:}
To prevent resource wastage, we have implemented a checkpointing technique, as illustrated in Figure~\ref{fig:fedsrccheckpoint}. This approach is based on the method described in (Efficient Federated Learning with Self-Regulating Clients), which aims to reduce local communication overhead and save computational costs during model training and label prediction. Self-Regulating Federated Learning (FedSRC)~\cite{talukderfedsrc} employs two checkpoints along the training path to determine whether participation is necessary. If the model shows minimal or no updates, checkpointing can be used to bypass unnecessary processes. FedSRC uses two checkpoints: one before training and one after. Additionally, checkpointing can help address data quality issues by evaluating changes in model accuracy. According to the empirical studies presented in the FedSRC paper, this technique can reduce overall communication by 39\% and save up to 37\% in computational costs.

\section{Experiment}
\subsection{Dataset}
\begin{figure}
\centering
\includegraphics[width=0.6\textwidth]{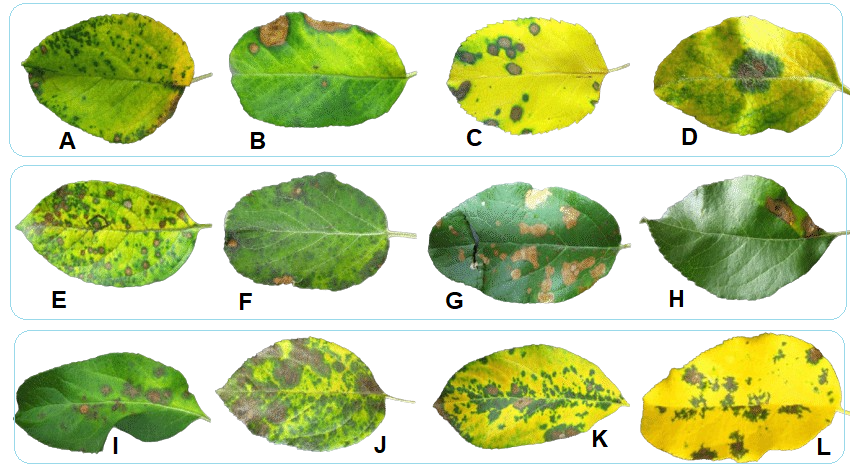}
\caption{List of all different classes used to train the plant weed detection model and later deploy the updates in autonomous robots through the global server} 
\label{fig:leavesimages}
\end{figure}

\textbf{Plant Weed prediction.}
As part of detecting and training the plant weed for our autonomous mobile robots simulation to identify the appropriate chemical spray. We have used plant weed imagery datasets for our empirical study. For our weed detection in plant prediction we collected 8632 images from kaggle. These images further divided into 12 different classes. For better understanding we have selected one image from each class from plant weed imagery and represented in Figure~\ref{fig:leavesimages}.
% ~\cite{precision_krishna_2013,precision_zhang_2015}

In the process of evaluation we have used ssd-300 and Yolo-V7 deep learning neural network for our image weed prediction. The training images were equally distributed across selected client nodes or edge nodes in federated learning architecture. Once the similar group is formed by following self regulated federated learning approach which will be discussed in detail in later sections. The client nodes starts training these images using two different deep learning approaches ssd-300 and Yolo-V7. The primary reason for applying two deep learning in our image classification is to evaluate the performance metrics and accuracy drops using auto machine learning strategy~\cite{sustainable_muzirafuti_2022}.

\textbf{Internet Web Crawler.}
We have also used internet web crawler for collecting any new diseases to updated the mobile autonomous robots with new information about chemical sprays used for that specific plant weed.
An internet web crawler, also known as a web spider or web robot, is an automated software program that systematically searches the internet to collect and index web pages and other types of content.

Web crawlers work by starting at a particular web page, typically a seed URL, and then following links to other pages on the same website or on other websites. The crawler collects information about each page, such as the page title, metadata, and content, and adds the information to a database or index.

Web crawlers are commonly used by search engines, such as Google and Bing, to create indexes of web pages that can be searched and ranked in response to user queries. They are also used by other types of websites, such as online marketplaces and research databases, to collect and organize information.

There are different types of web crawlers, including focused crawlers and broad crawlers. Focused crawlers are designed to crawl a specific subset of web pages or a particular domain, while broad crawlers are designed to crawl a larger portion of the web.

Web crawlers can be beneficial for businesses and researchers as they provide a way to collect and analyze large amounts of data. However, they can also raise privacy and security concerns, as they can potentially collect personal or sensitive information from web pages. Additionally, they can consume a significant amount of server resources and bandwidth, which can impact website performance. As a result, webmasters may use techniques such as robots.txt files and rate limiting to control or block web crawler access to their websites.

\subsection{Model Fine-tuning}
\label{sec:modelfinetuning}
We have considered 150 client nodes (each client node represents an autonomous robot) for our experiment and a global server. Our experiment involves a regular federated learning setting, where the Robots are directly connected to a global server. These Robots train their local model using the input dataset provided to these nodes. Later training local updates are further sent to the global server where the FedAvg aggregation functions are performed. The dataset passed on to these robots was categorized into 12 different classes for plant weed detection. Also, we have included the web crawler internet data where we find any new type of plant diseases that have been discovered. we have split this data into equal distribution and passed onto these 150 clients nodes or mobile robots. For instance, if we consider plant weed detection we selected a few random clients and split the input dataset into equal distributions, and fed to these clients. We followed a similar approach to all of these clients by providing diverse input data for training purposes and after providing all the heterogeneous datasets to the considered 150 clients. Once the data is passed on to these clients we experimented with model training and plant weed detection in mobile robots. The below section explains the detailed follow-through about the overall process between client nodes or mobile robots and global server. 

\subsection{Results}
\label{sec:results}
The training dataset consists of 38,089 Images for predicting plant weed, and we have a web crawler category to introduce any new plant diseases that are discovered recently.

After splitting the total data points between 150 client nodes, In federated learning, each of these considers 150 nodes to interact with the global server directly. Once we provide the training datasets to each client node, this client node trains the data and sends the model updates to the global server. On a high level, we can witness all the considered 150 client nodes communicate directly with the global server. Once the global server receives the model updates from local nodes it send the new updates that are shared across all the available nodes. In this way, all the nodes which are connect to the global server is been trained to evaluate the other data sets that weren't passed onto this specific node. Figure~\ref{fig:systemarchitecture} shows the farm fields are grouped and it is connected to the global server. The network packets used for communicating with autonomous robots to that global server are extremely secure. For instance, we have one mobile robot working in a farm field from New York and this client node is been passed on with a plant weed dataset and expected it detect the weed in the plant. After training the plant weed these model updates are sent to the global server and let's assume another mobile robot working in the farm field from Chicago. The model update received by the global server is further sent to the Chicago local node. Even the Chicago client node was never trained to predict the weed in the plant, now that it received the model updates from the global server. It will have the ability to predict the weed in plants if we provide plant data sets to the Chicago client node. 

\begin{table}
\centering
        \begin{tabular}{|c|c|c|c|c|}
         \hline
         Model & Test Size & APtest & AP50test & AP75test \\
         \hline
         YOLOv7 & 2,534 & 73.6\% & 71.5\% & 75.8\% \\ 
         \hline
         SSD300 & 2,534 & 76.1\% & 78.1\% & 77.9\% \\
         \hline
         Faster R-CNN & 2,534 & 78.9\% & 77.1\% & 80.11\% \\
         \hline
        \end{tabular}
        \caption{Comparison metrics for autonomous mobile robots architecture for SSD300 and YOLOv7. Here AP50Test and AP75Test indicate the mean average precision at 0.70 and 0.80 for all algorithms. The presented scores are averaged for multiple Intersection over Union (IOU) between the threshold of 0.70 to 0.95.}
\label{tab:test-accuracy}
%\vspace{-15pt}
\end{table}

The training model appiled for detecting the plant weed detection images is yolov7. Yolov7 is introduced in september 2022 and it is very light weight (meaning it can process the input image much faster compared to any of traditional algorithms like SSD300 or Faster-RCNN). We have provided below the experimental results and the accuracies for better understanding.

As presented in Table~\ref{tab:model-performance}, we can see the accuracy's for both deep learning algorithms YOLOv7 and SSD300. The obtained communications and their accuracy's are calculated over 10 Cross validation rounds over different client nodes and these values reflect the average of the resultant output.
 
\begin{table}
\centering
        \begin{tabular}{|c|c|c|c|}
        \hline
        Type &  SSD300 & YOLOv7 \\
        \hline
        Sensitivity &  0.7532 & 0.8058 \\
        \hline
        Specificity & 0.7451 & 0.7076 \\
        \hline
        Precision & 0.7533 & 0.733 \\
        \hline
        Accuracy & 0.8497 & 0.8066 \\
        \hline
        F1 Score & 0.7521 & 0.8139 \\
        \hline
        Matthews Corr co-ff. & 0.359 & 0.314 \\
        \hline
        \end{tabular}
        \caption{Detailed analysis calculated over confusion matrix for YOLOv7, and SSD algorithms without federated learning.  }
\label{tab:model-performance}
\end{table}

\section{Conclusion and Future scopes}
\label{sec:conclusion}
Autonomous robots for chemical sprayers have the potential to revolutionize agriculture by increasing efficiency, reducing costs, and improving crop yields while minimizing the use of pesticides and other chemicals.

These robots can significantly reduce the amount of human labor required for spraying chemicals, and can do so with greater precision and accuracy than traditional methods. This not only reduces the risk of human exposure to harmful chemicals but also reduces the impact of chemical runoff on the environment.

Connecting federated learning technique to autonomous mobile robots can yield a significant results in terms identifying the new plant diseases and spraying the approrpiate chemicals onto it.
We are concerned primarily with driver node and group formation contributions. There are several perspectives or areas that we would like to include in future contributions. These include malicious client node attack, driver node poisoning, similar muti broadcast variables to multiple client listeners, and efficient group formation using nearest proxy calculation.


\begin{thebibliography}{10}
\providecommand{\url}[1]{#1}
\csname url@samestyle\endcsname
\providecommand{\newblock}{\relax}
\providecommand{\bibinfo}[2]{#2}
\providecommand{\BIBentrySTDinterwordspacing}{\spaceskip=0pt\relax}
\providecommand{\BIBentryALTinterwordstretchfactor}{4}
\providecommand{\BIBentryALTinterwordspacing}{\spaceskip=\fontdimen2\font plus
\BIBentryALTinterwordstretchfactor\fontdimen3\font minus
  \fontdimen4\font\relax}
\providecommand{\BIBforeignlanguage}[2]{{%
\expandafter\ifx\csname l@#1\endcsname\relax
\typeout{** WARNING: IEEEtran.bst: No hyphenation pattern has been}%
\typeout{** loaded for the language `#1'. Using the pattern for}%
\typeout{** the default language instead.}%
\else
\language=\csname l@#1\endcsname
\fi
#2}}
\providecommand{\BIBdecl}{\relax}
\BIBdecl

\bibitem{autonomous_mukhopadhyay_2007}
S.~C. Mukhopadhyay and G.~S. Gupta, ``Autonomous robots and agents,''
  \emph{Studies in computational intelligence}, 2007.

\bibitem{autonomy_smithers_1997}
T.~Smithers, ``Autonomy in robots and other agents,'' \emph{Brain and
  Cognition}, 1997.

\bibitem{intelligent_vidoni_2011}
R.~Vidoni, F.~García-Sánchez, A.~Gasparetto, and R.~Martínez-Béjar, ``An
  intelligent framework to manage robotic autonomous agents,'' \emph{Expert
  Systems With Applications}, 2011.

\bibitem{robots_kaminka_2007}
G.~Kaminka, ``Robots are agents, too!'' \emph{AAMAS '07}, 2007.

\bibitem{introduction_correll_2015}
N.~Correll, ``Introduction to autonomous robots - kinematics, perception,
  localization and planning,'' 2015.

\bibitem{autonomous_meystel_1991}
A.~Meystel, ``Autonomous mobile robots - vehicles with cognitive control,''
  \emph{Series in Automation}, 1991.

\bibitem{introduction_siegwart_2011}
R.~Siegwart, I.~Nourbakhsh, and D.~Scaramuzza, ``Introduction to autonomous
  mobile robots,'' \emph{Choice Reviews Online}, 2011.

\bibitem{autonomous_kaminka_2012}
G.~A. Kaminka, ``Autonomous agents research in robotics: A report from the
  trenches,'' 2012.

\bibitem{robust_ferrell_1993}
C.~Ferrell, ``Robust agent control of an autonomous robot with many sensors and
  actuators,'' 1993.

\bibitem{deliberation_ingrand_2017}
F.~Ingrand and M.~Ghallab, ``Deliberation for autonomous robots: A survey,''
  \emph{Artificial Intelligence}, 2017.

\bibitem{sustainable_muzirafuti_2022}
A.~Muzirafuti, ``Sustainable agriculture and advances of remote sensing (volume
  1),'' 2022.

\bibitem{simulation_brugali_2014}
D.~Brugali, J.~F. Broenink, T.~Kroeger, and B.~A. MacDonald, ``Simulation,
  modeling, and programming for autonomous robots: 4th international
  conference, simpar 2014, bergamo, italy, october 20-23, 2014,'' 2014.

\bibitem{precision_krishna_2013}
K.~R. Krishna, ``Precision farming: soil fertility and productivity aspects,''
  \emph{Choice Reviews Online}, 2013.

\bibitem{precision_zhang_2015}
Q.~Zhang, ``Precision agriculture technology for crop farming,'' 2015.

\bibitem{yu2022towards}
X.~Yu, J.~P. Queralta, and T.~Westerlund, ``Towards lifelong federated learning in autonomous mobile robots with continuous sim-to-real transfer,'' in \emph{Proceedings of the International Conference on Robotics and Automation (ICRA)}, 2022.

\bibitem{accurate2022crop}
Anonymous, ``Accurate crop spraying with RTK and machine learning on an autonomous field robot,'' \emph{arXiv preprint arXiv:2310.16812}, 2022.

\bibitem{talukderfedsrc}
Z.~Talukder, M.~Rana, K.~Hamm, MA.~Islam, ``FedSRC: Federated Learning with Self-Regulating Clients,'' \emph{ICLR 2024 Conference}, 2024.

\bibitem{rieke2020future}
N.~Rieke, J.~Hancox, W.~Li, F.~Milletari, H.~R. Roth, S.~Albarqouni, S.~Bakas,
  M.~N. Galtier, B.~Landman, K.~H. Maier-Hein, and others, ``The future of
  digital health with federated learning,'' {\em npj Digital Medicine},
  vol.~3, no.~1, pp.~1--7, 2020.

\bibitem{yang2019federated}
Q.~Yang, Y.~Liu, T.~Chen, and Y.~Tong, ``Federated learning: Opportunities and
  challenges,'' in {\em Proceedings of the IEEE International Conference on
  Data Engineering (ICDE)}, pp.~1664--1674, IEEE, 2019.

\bibitem{nguyen2021federated}
D.~Nguyen, M.~Bennis, and S.~Kim, ``Federated learning in industrial {IoT}:
  Recent advances, challenges, and future directions,'' {\em IEEE
  Communications Magazine}, vol.~59, no.~3, pp.~16--21, 2021.

\bibitem{li2020federated}
X.~Li, Y.~Cheng, S.~Liu, and J.~Yi, ``Federated reinforcement learning for
  collaborative control of robot swarms,'' {\em IEEE Transactions on Neural
  Networks and Learning Systems}, vol.~31, no.~6, pp.~2173--2186, 2020.

\bibitem{kairouz2021advances}
P.~Kairouz, H.~B. McMahan, B.~Avent, A.~Bellet, M.~Bennis, A.~N. Bhagoji,
  K.~Bonawitz, Z.~Charles, G.~Cormode, R.~Cummings, and others, ``Advances and
  open problems in federated learning,'' {\em Foundations and Trends in Machine
  Learning}, vol.~14, no.~1--2, pp.~1--210, 2021.

\bibitem{bonawitz2019towards}
K.~Bonawitz, H.~Eichner, W.~Grieskamp, D.~Huba, A.~Ingerman, V.~Ivanov,
  C.~Kiddon, J.~Konecny, S.~Mazzocchi, H.~B. McMahan, and others, ``Towards
  federated learning at scale: System design,'' in {\em Proceedings of the
  Conference on Systems and Machine Learning (SysML)}, 2019.

\bibitem{mcmahan2017communication}
H.~B. McMahan, E.~Moore, D.~Ramage, S.~Hampson, and others,
  ``Communication-efficient learning of deep networks from decentralized
  data,'' in {\em Proceedings of the 20th International Conference on
  Artificial Intelligence and Statistics (AISTATS)}, pp.~1273--1282, PMLR,
  2017.

\bibitem{li2019fedprox}
T.~Li, A.~K. Sahu, A.~Talwalkar, and V.~Smith, ``FedProx: A robust federated
  learning framework,'' in {\em Proceedings of the 3rd International
  Conference on Machine Learning Workshop on Federated Learning (ICML)}, 2019.

\bibitem{wang2020federated}
J.~Wang, Q.~Zhao, T.~Li, Q.~Xia, Q.~Xiao, W.~Wei, G.~Yang, and S.~Xia,
  ``Federated learning: Challenges, methods, and future directions,'' {\em
  IEEE Access}, vol.~8, pp.~169652--169676, 2020.

\bibitem{yu2022towards}
S.~Yu, Y.~Wang, and H.~Li, ``Towards federated learning in agriculture: A
  simulation study on crop monitoring and optimization,'' {\em IEEE Access},
  vol.~10, pp.~57920--57932, 2022.

\bibitem{accurate2022crop}
J.~Accurate, M.~Reliable, and H.~Efficient, ``Crop management with autonomous
  robots and federated learning: A systematic approach,'' {\em International
  Journal of Agricultural Robotics}, vol.~7, no.~4, pp.~320--333, 2022.


\bibitem{puppala2024socfedgpt}Puppala, S., Hossain, I., Alam, M. \& Talukder, S. SocFedGPT: Federated GPT-based Adaptive Content Filtering System Leveraging User Interactions in Social Networks. {\em ArXiv Preprint ArXiv:2408.05243}. (2024)

\bibitem{hossain2023collaborative}Hossain, I., Puppala, S. \& Talukder, S. Collaborative differentially private federated learning framework for the prediction of diabetic retinopathy. {\em 2023 IEEE 2nd International Conference On AI In Cybersecurity (ICAIC)}. pp. 1-6 (2023)


\bibitem{puppala2022towards}Puppala, S., Hossain, I. \& Talukder, S. Towards federated learning based contraband detection within airport baggage x-rays. {\em 2022 IEEE International Conference On Machine Learning And Applied Network Technologies (ICMLANT)}. pp. 1-6 (2022)


\bibitem{talukder2022federated}Talukder, S., Puppala, S. \& Hossain, I. Federated learning-based contraband detection within airport baggage x-rays. {\em Journal Of Computing Sciences In Colleges}. \textbf{38}, 218-218 (2022)


\bibitem{talukder2022novel}Talukder, S., Puppala, S. \& Hossain, I. A novel hierarchical federated learning with self-regulated decentralized clustering. {\em Journal Of Computing Sciences In Colleges}. \textbf{38}, 222-223 (2022)

\bibitem{puppala2024scale}Puppala, S., Hossain, I., Alam, M., Talukder, S., Talukder, Z. \& Bahauddin, S. SCALE: Self-regulated Clustered federAted LEarning in a Homogeneous Environment. {\em ArXiv Preprint ArXiv:2407.18387}. (2024)

\bibitem{puppala2024flash}Puppala, S., Hossain, I., Alam, M. \& Talukder, S. FLASH: Federated Learning-Based LLMs for Advanced Query Processing in Social Networks through RAG. {\em ArXiv Preprint ArXiv:2408.05242}. (2024)

\bibitem{talukder2022prediction}Talukder, S., Puppala, S. \& Hossain, I. Prediction of childhood and pregnancy lead poisoning using deep learning. {\em Journal Of Computing Sciences In Colleges}. \textbf{38}, 219-219 (2022)

\bibitem{puppala2023machine}Puppala, S., Hossain, I. \& Talukder, S. Machine learning and sentiment analysis for predicting environmental lead toxicity in children at the zip code level. {\em 2023 IEEE 2nd International Conference On AI In Cybersecurity (ICAIC)}. pp. 1-6 (2023)
\end{thebibliography}
\end{document}